%% file: RFNN_camera_beta.tex
\documentclass[10pt,twocolumn,letterpaper]{article}
\usepackage{cvpr}
\usepackage{times}
\usepackage{epsfig}
\usepackage{graphicx}
\usepackage{amsmath}
\usepackage{amssymb}
\usepackage[export]{adjustbox}
\usepackage{tabularx}
\graphicspath{{images/}{./figures/}}
\usepackage{subfiles}
 
\usepackage{algorithmic}
\usepackage{algorithm}
\usepackage{subcaption}
\usepackage{float}
\usepackage{wrapfig}
\usepackage{booktabs}
\usepackage{epstopdf}

\usepackage[pagebackref=true,breaklinks=true,letterpaper=true,colorlinks,bookmarks=false]{hyperref}

\cvprfinalcopy % *** Uncomment this line for the final submission
\def\cvprPaperID{2543} % *** Enter the CVPR Paper ID here

\ifcvprfinal\fi

\makeatletter
\newcommand{\mypm}{\mathbin{\mathpalette\@mypm\relax}}
\newcommand{\@mypm}[2]{\ooalign{%
  \raisebox{.1\height}{$#1+$}\cr
  \smash{\raisebox{-.6\height}{$#1-$}}\cr}}
\makeatother

\begin{document}

%%%%%%%%% TITLE
\title{Structured Receptive Fields in CNNs}

\author{J{\"o}rn-Henrik Jacobsen$^1$, Jan van Gemert$^{1,2}$, Zhongyu Lou$^1$, Arnold W. M. Smeulders$^1$\\
$^1$University of Amsterdam, The Netherlands\\
$^2$TU Delft, The Netherlands\\
{\tt\small \{j.jacobsen,z.lou,a.w.m.smeulders\}@uva.nl, j.c.vangemert@tudelft.nl}}

\maketitle
%\thispagestyle{empty}

%%%%%%%%% ABSTRACT
\input{sections/abstract}

%%%%%%%%% BODY TEXT
\vspace{-5mm}
\subfile{sections/introduction}

%-------------------------------------------------------------------------

\subfile{sections/related_work}

\subfile{sections/method}

%-------------------------------------------------------------------------

\subfile{sections/experiments}

\subfile{sections/discussion}

%-------------------------------------------------------------------------

%\subfile{sections/conclusion}

{\small
\bibliographystyle{ieee}
\bibliography{RFNN_camera_beta}
}

\end{document}

%% file: sections/abstract.tex
 \begin{abstract}
Learning powerful feature representations with CNNs is hard when training data are limited. Pre-training is one way to overcome this, but it requires large datasets sufficiently similar to the target domain. Another option is to design priors into the model, which can range from tuned hyperparameters to fully engineered representations like Scattering Networks. We combine these ideas into structured receptive field networks, a model which has a fixed filter basis and yet retains the flexibility of CNNs. This flexibility is achieved by expressing receptive fields in CNNs as a weighted sum over a fixed basis which is similar in spirit to Scattering Networks. The key difference is that we learn arbitrary effective filter sets from the basis rather than modeling the filters. This approach explicitly connects classical multiscale image analysis with general CNNs. With structured receptive field networks, we improve considerably over unstructured CNNs for small and medium dataset scenarios as well as over Scattering for large datasets. We validate our findings on ILSVRC2012, Cifar-10, Cifar-100 and MNIST. As a realistic small dataset example, we show state-of-the-art classification results on popular 3D MRI brain-disease datasets where pre-training is difficult due to a lack of large public datasets in a similar domain.
 \end{abstract}

%% file: sections/introduction.tex
\section{Introduction}

Where convolutional networks have appeared enormously powerful in the classification of images when ample data are available~\cite{lecunNature15deep}, we focus on smaller image datasets. We propose structuring receptive fields in CNNs as linear combinations of basis functions to train them with fewer image data.

\begin{figure}
\includegraphics[width=.48\textwidth, center]{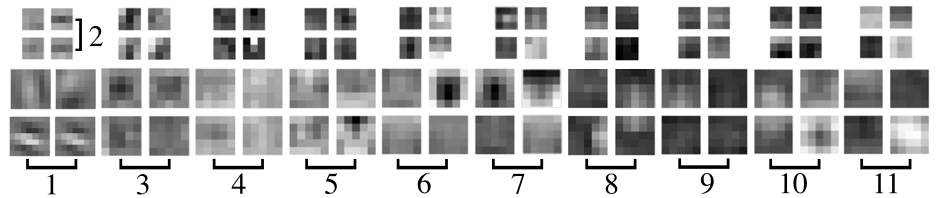}
\vspace{-5mm}
\caption{Filters randomly sampled from all layers of the GoogLenet model \cite{szegedyArxiv14goingDeeperGoogLEnet}, from left to right layer number increases. Without being forced to do so, the model exhibits spatial coherence (seen as smooth functions almost everywhere) after being trained on ILSVRC2012. This behaviour reflects the spatial coherence of the input feature maps even in the highest layers.  \label{fig:overflow}}
\end{figure}

The common approach to smaller datasets is to perform pre-training on a large dataset, usually ImageNet~\cite{russakovskyIJCV14imagenet}. Where CNNs generalize well to domains similar to the domain where the pre-training came from~\cite{razavianCVPRw14cnnOffTheShelf,zhouNIPS14placesDataset}, the performance decreases significantly when moving away from the pre-training domain~\cite{zhouNIPS14placesDataset,yosinski2014transferable}. We aim to make learning more effective for smaller sets by restricting CNNs parameter spaces. Since \emph{all images} are spatially coherent and human observers are considered to only cast local variations up to a certain order as meaningful~\cite{koenderinkBioCyb4structOfIm,lindebergBook13scaleSpaceInCV} our key assumption is that it is unnecessary to learn these properties in the network. When visualizing the intermediate layers of a trained network, see e.g.~\cite{zeilerECCV14visualizingCNN} and Figure~\ref{fig:overflow}, it becomes evident that the filters as learned in a CNN are locally coherent and as a consequence can be decomposed into a smooth compact filter basis~\cite{koenderinkBioCyb87localGeoInVisSys}.

\begin{figure}
\centering
\includegraphics[width=.48\textwidth]{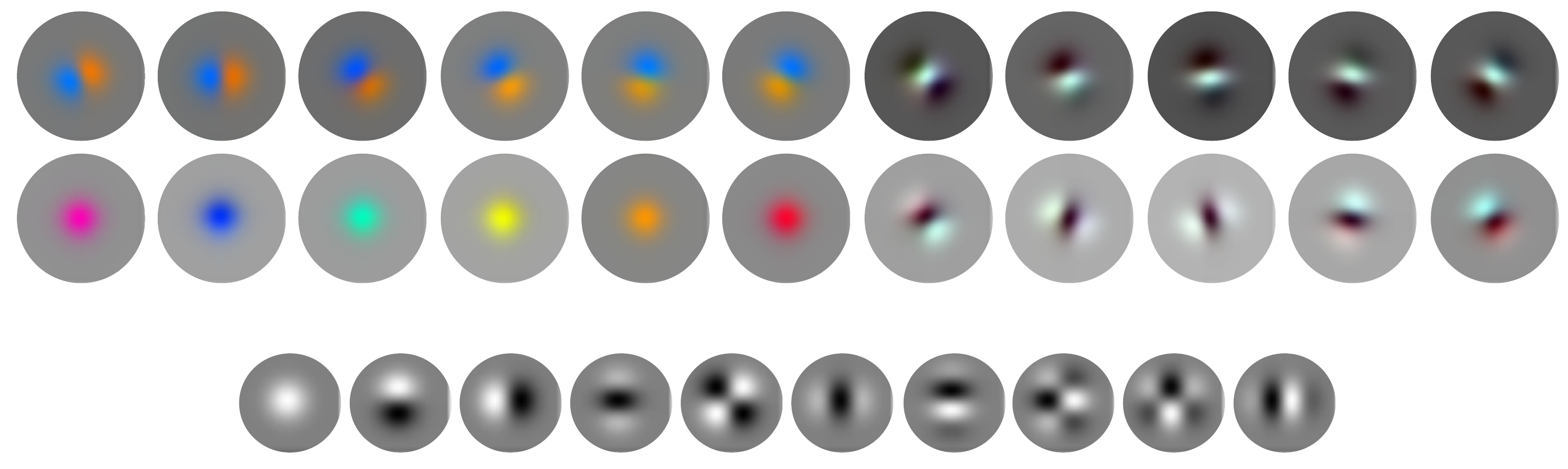}
\vspace{-5mm}
\caption{ 
A subset of filters of the first structured receptive field CNN layer as trained on 100-class ILSVRC2012 and the Gaussian derivative basis they are learned from. The network learns scaled and rotated versions of zero, first, second and third order filters. Furthermore, the filters learn to recombine the different input color channels which is a crucial property of CNNs. %The filters are ordered from left to right by dominating basis order and rotation angle, showing that rotated first and second order derivatives dominate the first layer.
}
\label{fig:visLayers}
\end{figure}

We aim to maintain the CNN's capacity to learn general variances and invariances in arbitrary images. Following from our assumptions, the demand is posed on the filter set that i) a linear combination of a finite basis set is capable of forming any arbitrary filter necessary for the task at hand, as illustrated in Figure~\ref{fig:visLayers} and ii) that we preserve the full learning capacity of the network. For i) we choose the family of Gaussian filters and its smooth derivatives for which it has been proven~\cite{koenderinkBioCyb87localGeoInVisSys} that 3-rd or 4-th order is sufficient to capture all local image variation perceivable by humans. According to scale-space theory~\cite{koenderinkBioCyb4structOfIm,WitkinIJCAI83scaleSpace}, the Gaussian family constitutes the Taylor expansion of the image function which guarantees completeness. For ii) we maintain backpropagation parameter optimization in the network, now applied to learning the weights by which the filters are summed into the effective filter set.

Similarly motivated, the Scattering Transform~\cite{brunaTPAMI13scattering,mallat2012groupInvScat,sifreCVPR13rotScat}, a special type of CNN, uses a complete set of wavelet filters ordered in a cascade. However, different from a classical CNN, the filters parameters are not learned by backpropagation but rather they are fixed from the start and the whole network structure is motivated by signal processing principles. In the Scattering Network the choice of local and global invariances are tailored to the type of images specifically. In the Scattering Transform invariance to group actions beyond local translation and deformation requires explicit design~\cite{mallat2012groupInvScat} with the regards to the variability encountered in the target domain such as translation~\cite{brunaTPAMI13scattering}, rotation~\cite{sifreCVPR13rotScat} or scale. As a consequence, when the desired invariance groups are known a priori, Scattering delivers very effective networks. \\
Our paper takes the best of two worlds. On the one hand, we adopt the Scattering principle of using fixed filter bases as a function prior in the network. But on the other hand, we maintain from plain CNNs the capacity to learn arbitrary effective filter combinations to form complex invariances and equivariances.

% Comparison of SRFNNs and CNNs on a 100-class subset of ILSVRC2012 results in a performance superior to the Network in Network~\cite{linArxiv13NetwInNetw} and Alexnet architecture~\cite{krizhevskyNIPS12imagenet}.

Our main contributions are:
\begin{itemize}
\item Deriving the structured receptive field network (RFNN) from first principles by formulating filter learning as a linear decomposition onto a filter basis, unifying CNNs and multiscale image analysis in a learnable model.
\item  Combining the strengths of Scattering and CNNs. We do well on both domains: i) small datasets where Scattering is best but CNNs are weak; ii) complex datasets where CNNs excel but Scattering is weak. 
\item State-of-the-art classification results on a small dataset where pre-training is infeasible. The task is Alzheimer's disease classification on two widely used brain MRI datasets. We outperform all published results on the ADNI dataset.
\end{itemize}

%% file: sections/related_work.tex
\section{Related Work}

%\subsection{Structured Receptive Fields}

\subsection{Scale-space: the deep structure of images}

Scale-space theory~\cite{WitkinIJCAI83scaleSpace} provides a model for the structure of images by steadily convolving the image with filters of increasing scale, effectively reducing the resolution in each scale step. While details of the image will slowly disappear, the order by which they do so will uniquely encode the deep structure of the image~\cite{koenderinkBioCyb4structOfIm}. Gaussian filters have the advantage in that they do not introduce any artifacts~\cite{lindebergBook13scaleSpaceInCV} in the image while Gaussian derivative filters form a complete and stable basis to decompose locally any realistic image. The set of responses to the derivative filters describing one patch is called the N-jet~\cite{florackIVC92scaleAndDiffStruct}.
 
In the same vein, CNNs can be perceived to also model the deep structure of images, this time in a non-linear fashion. The pooling layers in a CNN effectively reduce resolution of input feature maps. Viewed from the top of the network down, the spatial extent of a convolution kernel is increased in each layer by a factor 2, where a 5x5 kernel at the higher layer measures 10x10 pixels on the layer below. The deep structure in a CNN models the image on several discrete levels of resolution simultaneously, precisely in line with Scale-space theory. 
 
Where CNNs typically reduce resolution by max pooling in a non-linear fashion, Scale-space offers a linear theory for continuous reduction of resolution. Scale-space theory treats an image as a function of the mathematical apparatus to reveal the local image structure. In this paper, we exploit the descriptive power of Scale-space theory to decompose the image locally on a fixed filter basis of multiple scales.

%Scale space theory~\cite{florackIVC92scaleAndDiffStruct,koenderinkBioCyb4structOfIm,lindebergBook13scaleSpaceInCV,WitkinIJCAI83scaleSpace} and CNNs both model the deep structure of images. The pooling layers in a CNN subsample a feature map to half of its resolution. In effect, the spatial extent of a convolution kernel becomes 4 times larger: a 5x5 kernel at the higher layer measures 20x20 pixels on the layer below. The deep structure in a CNN models the image on several discrete levels of resolution simultaneously, which is exactly what scale space theory prescribes. Where CNNs are hard-coded to sub-sample in factors of 2, scale space offers a rich mathematical theory for multi-resolution modeling. The main difference is that CNNs treat an image as an array of pixels whereas scale space treats an image as a function, offering fundamental mathematical operations like differentiations, to reveal local image structure. In this paper we exploit the benefits of scale space theory to add structure to receptive field convolution filters in CNNs.

\subsection{CNNs and their parameters}

CNNs~\cite{lecunIEEE1998CNNs} have large numbers of parameters to learn~\cite{krizhevskyNIPS12imagenet}. This is their strength as they can solve extremely complicated problems~\cite{krizhevskyNIPS12imagenet,taigmanCVPR14deepface}. At the same time, their number of unrestricted parameters is a limiting factor in terms of the large amounts of data needed to train. To prevent overfitting, which is an issue even when training on large datasets like the million images of the ILSVRC2012 challenge~\cite{russakovskyIJCV14imagenet}, usually regularization is imposed with methods like dropout~\cite{srivastavaJMLR2014dropout} and weight decay~\cite{moodyNIPS1995weightdecay}. Regularization is essential to achieving good performance. In cases where limited training data are available, CNN training quickly overfits regardless and the learned representations do not generalize well. Transfer learning from models pre-trained in similar domains to the new domain is necessary to achieve competitive results~\cite{OquabCVPR14transferCNNfeatures}. One thing pre-training on large datasets provides is knowledge about properties inherent to all natural images, such as spatial coherence and robustness to uninformative variability. In this paper, we aim to design these properties into CNNs to improve generalization when limited training data are available.

\subsection{The Scattering representation}
To reduce model complexity we draw inspiration from the elegant convolutional Scattering Network~\cite{brunaTPAMI13scattering,mallat2012groupInvScat,sifreCVPR13rotScat}. Scattering uses a multi-layer cascade of a pre-defined wavelet filter bank with nonlinearity and pooling operators. It computes a locally translation-invariant image representation, stable to deformations while avoiding information loss by recovering wavelet coefficients in successive layers. No learning is used in the image representation: all relevant combinations of the filters are fed into an SVM-classifier yielding state-of-the-art results on small dataset classification. Scattering is particularly well-suited to small datasets because it refrains from feature learning. Since all filter combinations are pre-defined, their effectiveness is independent of dataset size. In this paper, we also benefit from a fixed filter bank. In contrast to Scattering, we \emph{learn} linear combinations of a filter basis into effective filters and non-linear combinations thereof. 

The wavelet filterbank of Scattering is carefully designed to sample a range of rotations and scales. These filters and their properties are grounded in wavelet theory~\cite{mallat1999wavelet} and exhibit precisely formulated properties. By using interpretable filters, Scattering can design invariance to finite groups such as translation~\cite{brunaTPAMI13scattering}, scale and rotation~\cite{sifreCVPR13rotScat}. Hard coding the invariance into the network is effective when the problem and its invariants are known precisely, but for many applications this is rarely the case. When the variability is unknown, additional Scattering paths have to be computed, stored and processed exhaustively before classification. This leads to a well-structured but very high dimensional parameter space. In this paper, we use a Gaussian derivatives basis as the filter bank, firmly grounded in scale-space theory~\cite{koenderinkBioCyb4structOfIm,lindebergBook13scaleSpaceInCV,WitkinIJCAI83scaleSpace}. Our approach incorporates learning effective filter combinations from the very beginning, which allows for a compact representation of the problem at hand.

\subsection{Recent CNNs}
% Since the breakthrough of Alexnet~\cite{krizhevskyNIPS12imagenet}, there has been much effort to improve on the reported results in Krizhevsky's 2012 paper. Impressive jumps in performance~\cite{linArxiv13NetwInNetw,simonyanArxiv14veryDeepVGG,szegedyArxiv14goingDeeperGoogLEnet} have been achieved by sophisticated parameter reduction and different ways to regularize the models while retaining their expressiveness. 

Restriction of parameter spaces has led to some major advances in recent CNNs performance. Network in Network \cite{linArxiv13NetwInNetw} and GoogleNet \cite{szegedyArxiv14goingDeeperGoogLEnet} illustrate that fully connected layers, which constitute most of Alexnet's parameters, can be replaced by a global average pooling layer reducing the number of parameters in the fully connected layers to virtually zero. The number of parameters in the convolution layers is increased to enhance the expressiveness of each layers features. Overall the total number of parameters is not necessarily decreased, but the function space is restricted, allowing for bigger models while classification accuracy improves~\cite{linArxiv13NetwInNetw,szegedyArxiv14goingDeeperGoogLEnet}. 

The VGG Network~\cite{simonyanArxiv14veryDeepVGG} improves over Alexnet in a different way. The convolution layers parameter spaces are restricted by splitting each 5x5 convolution layer into two 3x3 convolution layers. 5x5 convolutions and 2 subsequent 3x3 convolutions have the same effective receptive field size while each receptive field has 18 instead of 25 trainable parameters. This regularization enables learning larger models that are less prone to overfitting. In this paper, we follow a different approach in restricting the free parameter space without reducing filter size.

%% file: sections/method.tex
\section{Deep Receptive Field Networks}
\label{sect:method}

\subsection{Structured receptive fields}

%Scale space offers an unbiased front-end for a vision system~\cite{florackIVC92scaleAndDiffStruct,koenderinkBioCyb87localGeoInVisSys}. Its sole task is to provide a representation of the visual information without committing to any interpretation. The requirements for an unbiased front end~\cite{florackIVC92scaleAndDiffStruct} are: linearity, allow for input superposition; Spatial shift invariance, no preferred location; isotropy, no preferred direction; scale invariance, no preferred scale.

%
%The front-end decrease the burden on interpretation

%representation only, no interpretation.
%
%- linearity
%- spatial shift invariant
%- isotropy
%- scale invariant
%
% decrease the burden on interpretation, less overhead, known in advance, the front end system will make this a priori knowledge of the environment manifets

%- Image is a function, not as pixels. \\
%- kernel is an aperture, receptive field \\

In our structured receptive field networks we make the relationship between Scale-space and CNNs explicit. Whereas normal CNNs treat images and their filters as pixel values, we aim for a CNN that treats images as functions in Scale-space. Thus, the learned convolution kernels become functions as well. We therefore approximate an arbitrary CNN filter $F(x)$ with a Taylor expansion around $a$ up to order $M$
\begin{equation}
F(x) = \sum^M_{m=0} \frac{F^m(a)}{m!} (x - a)^m.
\end{equation}
Scale-space allows us to use differential operators on images, due to linearity of convolution we are able to compute the exact derivatives of the scaled underlying function by convolution with derivatives of the Gaussian kernel
\begin{equation}\label{eq:taylor}
G(.;\sigma)\ast F(x) = \sum^N_{m=0} \frac{(G^m(.;\sigma)\ast F)(a)}{m!} (x - a)^m ,
\end{equation}
where $\ast$ denotes convolution, $G(.;\sigma)$ is a Gaussian kernel with scale $\sigma$ and $G^m(.;\sigma)$ is the $m^{th}$ order Gaussian derivative with respect to it's spatial variable. Thus, a convolution with a basis of weighted Gaussian derivatives receptive fields is the functional equivalent to pixel values in a standard CNN operating on a scaled infinitely differentiable version of the image. \par
To construct the full basis set in practice, one can show that the Hermite polynomials emerge from a sequence of Gaussian derivatives up to order $M$~\cite{romeny2008front}. A Gaussian derivative of arbitrary order can be obtained from the orthogonal Hermite polynomials $H_m$ through pointwise multiplication with a Gaussian envelope
\begin{equation}
\label{eq:hermite}
G^m(.;\sigma) =(-1)^m \frac{1}{\sqrt{\sigma}^m} H_m(\frac{x}{\sigma\sqrt{2}}) \circ G(x;\sigma).
\end{equation}
The resulting operators allow computation of an image's local geometry at scale $\sigma$ and location $x$ up to any order of precision $M$. This basis is thus a complete set. Each derivative corresponds to an independent degree of freedom, making it also a minimal set. \\
Thus, an RFNN is a general CNN when a complete polynomial up to infinite order is considered. We restrict the basis based on the requirement that one can construct quadrature pair filters as suggested by Scattering and by evidence from Scale-space theory~\cite{koenderinkBioCyb87localGeoInVisSys} that considers all orders up to a maximum of 4, as it has been suggested that orders beyond that do not carry any information meaningful to visual perception.\\

\subsection{Transformation properties of the basis}

The isotropic Gaussian derivatives exhibit multiple desirable properties. It is possible to create complex multi-orientation pyramids that constitute wavelet representations similar to the Morlet Wavelet pyramids used in Scattering Networks~\cite{brunaTPAMI13scattering}. A complex multiresolution filterbank can be constructed from a dilated and rotated Gaussian derivative quadrature. The exact dilated versions of an arbitrary Gaussian derivative $G^m$ can be obtained through convolution with a Gaussian kernel of scale $\sigma=n$ according to
\begin{equation}
G^m(.;\sqrt{j^2+n^2})=G^m(.;j) \ast G(.;n).
\end{equation}
Arbitrary rotations of Gaussian derivative kernels can be obtained from a minimal set of basis filters without the need to rotate the basis itself. This property is referred to as steerability~\cite{freemanTPAMI91steerable}. Steerability is a property of all functions that can be expressed in a polynomial in x and y times an isotropic Gaussian. This certainly holds for the Gaussian derivatives according to equation \ref{eq:hermite}. For example a quadrature pair of $2^{\text{nd}}$ and $3^{\text{rd}}$ order Gaussian derivatives $G^{xx}$  and $G^{xxx}$ rotated by an angle $\theta$ can be obtained from a minimal 3 and 4 x-y separable basis set given by

\begin{equation}
\begin{split}
G^{xx}_{\theta}=\cos^2(\theta) G^{xx} - 2 \cos(\theta)\sin(\theta) G^{xy} + \sin^2(\theta) G^{yy} \\
G^{xxx}_{\theta}=\cos^3(\theta) G^{xxx} - 3 \cos^2(\theta) \sin(\theta) G^{xxy} \\ 
+ 3 \cos(\theta) \sin^2(\theta) G^{xyy} - \sin^3(\theta) G^{yyy}
\end{split}
\end{equation} 
A general derivation of the minimal basis set necessary for steering arbitrary orders can be found in~\cite{freemanTPAMI91steerable}. Note that the anisotropic case can be constructed in analogous manner according to~\cite{perona1992steerable}. This renders Scattering as a special case of the RFNN for fixed angles and scales, given a proper choice of pooling operations and possibly skip connections to closely resemble the architecture described in~\cite{brunaTPAMI13scattering}. In practice this allows for seamless integration of the Scattering concept into CNNs to achieve a variety of hybrid architectures.

\subsection{Learning basis filter parameters}

\begin{algorithm}[t]
\caption{RFNN Learning - updating the parameters $\alpha^{l}_{ij}$ between input map indexed by $i$ and output map indexed by $j$ of layer $l$ in the Mini-batch Gradient Decent framework.}
\label{algorithm1}
\begin{algorithmic}[1]
  \STATE {\bfseries Input:} input feature maps $o^{l-1}_{i}$ for each training sample (computed for the previous layer, $o^{l-1}$ is the input image when $l=1$), corresponding ground-truth labels $\{{y}_1,{y}_2,\dots,{y}_K\}$, the basic kernels $\{{\phi}_1,{\phi}_2,\dots,{\phi}_M\}$, previous parameter $\alpha^{l}_{ij}$.
  %\\\hrulefill
   \STATE compute the convolution $\{\zeta_1,\zeta_2,\dots,\zeta_m\}$ of $\{{o^{l-1}}_i\}$ respect to the basic kernels $\{{\phi}_1,{\phi}_2,\dots,{\phi}_M\}$
   \STATE obtain the output map $o^l_{j} = \alpha^{l}_{ij1}\cdot\zeta_1+\alpha^{l}_{ij2}\cdot\zeta_2+...+ \alpha^{l}_{ijM}\cdot\zeta_M$
   \STATE compute the $\delta^l_{jn}$ for each output neuron $n$ of the output map $o^l_{j}$ 
   \STATE compute the derivative $\psi'(t^l_{jn})$ of the activation function 
   \STATE compute the gradient $\frac{\partial E}{\partial \alpha^l_{ij}}$ respect to the weights $\alpha^{l}_{ij}$ 
   \STATE update parameter  $\alpha^l_{ij}=\alpha^l_{ij} - r\cdot \frac{1}{K}\cdot\sum_{k=1}^K [\frac{\partial E}{\partial \alpha^l_{ij}}]_{k}$, $r$ is the learning rate
   \STATE {\bfseries Output:} $\alpha^{l}_{ij}$, the output feature maps $o^{l}_{j}$
\end{algorithmic}
\end{algorithm}

%CNNs are typically \jvg{'typically'? so not always? Why is this important?} trained with the backpropagation algorithm \cite{lecunNature15deep} \jvg{for whom is that first sentence? Remove?}. 

%For convolutional neural networks, the weights to be learned are the pixels of the filter kernels \jvg{why relevant?}. Traditionally, the filter kernels are randomly initialized [REF] \jvg{add missing ref} and updated in a stochastic gradient descend. In our approach, the parameters of the convolutional layers are the parameters $\alpha$, see equation \ref{eq:filter}. These $\alpha$ are learned in a mini-batch gradient descent framework. \jvg{good!} \jvg{but why did the initialization disappear?} \jvg{this can be shorter}

Learning a feature representation boils down to convolution kernel learning. Where a classical CNN learns pixel values of the convolutional kernel, a RFNN learns Gaussian derivative basis function weights that combine to a convolution kernel function. A 2D filter kernel function $F(x,y)$ in all layers, is a linear combination of $i$ unique (non-symmetric) Gaussian derivative basis functions $\phi$
\begin{equation}
\label{eq:filters}
F(x,y) = \alpha_1 \phi_1 + \cdots + \alpha_n \phi_i,
\end{equation}
where  $\alpha_1,...,\alpha_i$ are the parameters being learned. 

\begin{figure}
\includegraphics[width=0.4\textwidth, center]{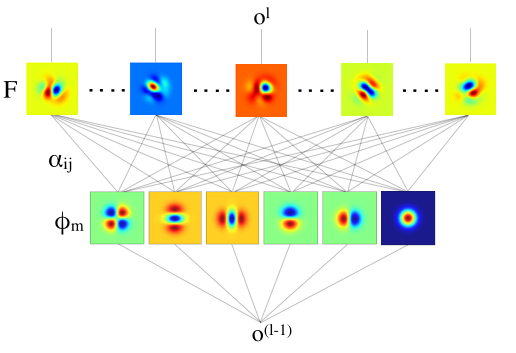}
\caption{An illustration of the basic building block in an RFNN network. A linear comibination of a limited basis filter set $\phi_{m}$ yields an arbitrary number of effective filters. The weights $\alpha_{ij}$ are learned by the network. 
\label{fig:basislayer}}
\end{figure}

We learn the filter's weights $\alpha$ by mini-batch stochastic gradient descent and compute the derivatives of the loss function $E$ with respect to the parameters $\alpha$ through backpropagation. It is straightforward to show the independence between the basis weights $\alpha$ and the actual basis (see Appendix for derivation). Thus, we formulate the basis learning as a combination of a fixed basis layer with a 1x1 convolution layer that has a kernel depth equal to the basis order. Propagation through the 1x1 layer is done as in any CNN while propagation through the basis layer is achieved by a convolution with flipped versions of the Gaussian filters. This makes it straightforward to include into any existing deep learning framework. The basic structured receptive field building block is illustrated in figure \ref{fig:basislayer}, showing how each effective filter is composed out of multiple basis filters. Note that the linearity of convolution allows us to never actually compute the effective filters. Convolving with effective filters is the same as convolving with the basis and then recombining the feature maps, allowing for efficient implementation. Algorithm \ref{algorithm1} shows how the parameters are updated.

\subsection{The network}

In this work, we choose the Network in Network (NiN) architecture~\cite{linArxiv13NetwInNetw} as the basis into which we integrate the structured receptive fields. It is particularly suited for an analysis of the RFNN approach, as the absence of a fully connected layer ensures all parameters to be fully concerned with re-combining basis filter outcomes of the current layer. At the same time, it is powerful, similar in spirit to the state of the art Googlenet~\cite{szegedyArxiv14goingDeeperGoogLEnet}, while being comparably small and fast to train. 

NiN alternates one spatial convolution layer with 1x1 convolutions and pooling. The 1x1 layers form non-linear combinations of the spatial convolution layers outputs. This procedure is repeated four times in 16 layers, with different number of filters and kernel sizes for the spatial convolution layer. The final pooling layer is a global average pooling layer. Each convolution layer is followed by a rectifier nonlinearity. Details on the different NiNs for Cifar and Imagenet can be found in the Caffe model zoo~\cite{jia2014caffe}. 

In the RFNN version of the Network in Network model, the basis layer including the Gaussian derivatives set is replacing the spatial convolution layer and corresponds to $\phi_m$ in equation \ref{eq:filters}. Thus, each basis convolution layer has a number of filters depending on order and scale of the chosen basis set. The basis set is fixed: no parameters are learned in this layer. The linear re-combination of the filter basis is done by the subsequent 1x1 convolution layer, corresponding to $\alpha_{ij}$ in equation \ref{eq:filters}. Note that there is no non-linearity between $\phi_m$ and $\alpha_{ij}$ layer in the RFNN case, as the combinations of the filters are linear. Thus the RFNN model is almost identical to the standard Network in Network. \\
We evaluate the model with and without multiple scales $\sigma_s$. When including scale, we extract 4 scales, as the original model includes 3 pooling steps and thus operates on 4 scales at least. In the first layer we directly compute 4 scales, sampled continuously with $\sigma_s=2^{s}$ where $s=scale$ as done in~\cite{brunaTPAMI13scattering}. In each subsequent layer we discard the lowest scale. The dimensionality reduction by max pooling renders it meaningless to insert the lowest scale of the previous layer into the filter basis set as it is already covered by the pyramidal structure of the network. This enables us to save on basis filters in the higher layers of the network. In conclusion we reduce the total number of 2D filters in the network from 520,000 in the standard Network in Network to between 12 and 144 in the RF Network in Network (RFNiN), while retaining the models expressiveness as shown in the experimental section.

%% file: sections/experiments.tex
\section{Experiments}

The experiments are partitioned into four parts. i) We show insight in the proposed model to investigate design choices; ii) we show that our model combines the strengths of Scattering and CNNs; iii) we show structured receptive fields improve classification performance when limiting training data; iv) we show a 3D version of our model that outperforms the state-of-the-art, including a 3D-CNN, on two brain MRI classification datasets where large pre-training datasets are not available. We use the Caffe library~\cite{jia2014caffe} and Theano~\cite{bastien2012theano} where we added RFNN as a separate module. Code is available on github\footnote{https://github.com/jhjacobsen/RFNN}.

\subsection{Experiment 1: Model insight}

The RFNN used in this section is the structured receptive field version of the Network in Network (RFNiN) introduced in section 3.3. We gain insight into the model by evaluating the scale and order of the basis filters. In addition, we analyze the performance compared to the standard Network in Network (NiN)~\cite{linArxiv13NetwInNetw} and Alexnet~\cite{krizhevskyNIPS12imagenet} and show that our proposed model is not merely a change in architecture. To allow overnight experiments we use the 100 largest classes of the ILSVRC2012 ImageNet classification challenge~\cite{russakovskyIJCV14imagenet}. Selection is done by folder size, as more than 100 classes have 1,300 images in them, yielding a dataset size of 130,000 images. This is a real-world medium sized dataset in a domain where CNNs excel. 

\textbf{Experimental setup}. The Network in Network (NiN) model and our Structured Receptive Field Network in Network (RFNiN) model are based on the training definitions provided by the Caffe model zoo~\cite{jia2014caffe}. Training is done with the standard procedure on Imagenet. We use stochastic gradient descent, a momentum of 0.9, a weight decay of 0.0005. The images are re-sized to 256x256, mirrored during training and the dataset mean is subtracted. The base learning rate was decreased by a factor of 10, according to the reduction from 1,000 to 100 classes, to ensure proper scaling of the weight updates, NiN didn't converge with the original learning rate. We decreased it by a factor of 10 after 50,000 iterations and again by the same factor after 75,000 iterations. The networks were trained for 100,000 iterations. Results are computed as the mean Top-1 classification accuracy on the validation set. \\

\begin{table}
\begin{center}
\begin{tabular}{@{}llll@{}} 
\toprule
\multicolumn{4}{l}{ILSVRC2012-100 Subset}\\ \midrule

Method & Top-1 & 2DFilters & $\#$Params \\ \midrule    %& Accuracy LR 0.01 \\

RFNiN $1^{st}$-order & 44.83\%  &$12$ & 1.8M \\
RFNiN $2^{nd}$-order & 61.24\%  &$24$ & 3.4M \\
RFNiN $3^{rd}$-order & 63.64\%  &$40$ & 5.5M\\
RFNiN $4^{th}$-order & 62.92\%  &$60$ & 8.1M\\
RFNiN-Scale $1^{st}$-order &  57.21\% &$24$ & 2.2M\\
RFNiN-Scale $2^{nd}$-order &  67.56\% &$54$& 4.2M\\
RFNiN-Scale $3^{rd}$-order &  \textbf{69.65\%} &$94$ & 6.8M\\ 
RFNiN-Scale $4^{th}$-order &  68.95\% &$144$   & 10.1M\\
\midrule   
Network in Network & 67.30\% &$520k$ &  8.2M\\
Alexnet & 54.86\% &$370k$ & 60.0M\\  
\bottomrule
%\hline\hline
%Alexnet \textit{Pre-train} & 71.99 / 64.88  \% &$3,689,28$ & \\
%Network in Network \textit{Pre-train} &  /  64.01 \% &$516,384$ &  \\
\end{tabular}
\end{center}
\vspace{-5mm}
\caption{Results on 100 Biggest ILSVRC2012 classes: The table shows RFNiN with 1st, 2nd, 3rd and 4th order filters in the whole network. Row 1-4 are applying basis filters in all layers on a scale of $\sigma$=1. RFNiN-scale in row 5-8 applies basis filters on 4 scales, where $\sigma$=1,2,4,8. The results show that a $3^{rd}$ order basis is sufficient while incorporating scale into the network gives a big gain in performance. The RFNiN is able to outperform the same Network in Network architecture.}
\label{tab:ILSVRC100}
\end{table}

\textbf{Filter basis order}. 
In table~\ref{tab:ILSVRC100}, the first four rows show the result of RFNiN architectures with 1st to 4th order Gaussian derivative basis filter set comprised of 12 to 60 individual Gaussian derivative filters in all layers of the network. In these experiments the value of $\sigma$=1, fixed for all filters and all layers. Comparing first to fourth order filter basis in table~\ref{tab:ILSVRC100}, we conclude that third order is sufficient, outperforming first and second order as predicted by Scale-space theory~\cite{koenderinkBioCyb87localGeoInVisSys}. The fourth order does not add any more gain.

\textbf{Filter scale}. The RFNiN-Scale entries of table~\ref{tab:ILSVRC100} show the classification result up to fourth order now with 4 different scales, $\sigma$=1, 2, 4, 8 for the lowest layer, $\sigma$=1, 2, 4 for the second layer, $\sigma$=1, 2 for the third, and $\sigma$=1 for the fourth. This implies that the basis filter set expands from 24 up to 144 filters in total in the network. Comparing the use of single scale filters in the network to dilated copies of the filters with varying scale indicates that a considerable gain can be achieved by including filters with different scales. This observation is supported by Scattering~\cite{brunaTPAMI13scattering}, showing that the multiple scales can directly be extracted from the first layer on. In fact, normal CNNs are also capable of similar behavior, as positive valued low-pass filter feature maps are not affected by rectifier nonlinearities~\cite{sifreCVPR13rotScat}. Thus, scale can directly be computed from the first layer onwards, which yields a much smaller set of basis filters and fewer convolutions needed in the higher layers. Note that number of parameters is not directly correlated with performance.

\begin{figure}
\includegraphics[width=.45\textwidth, left]{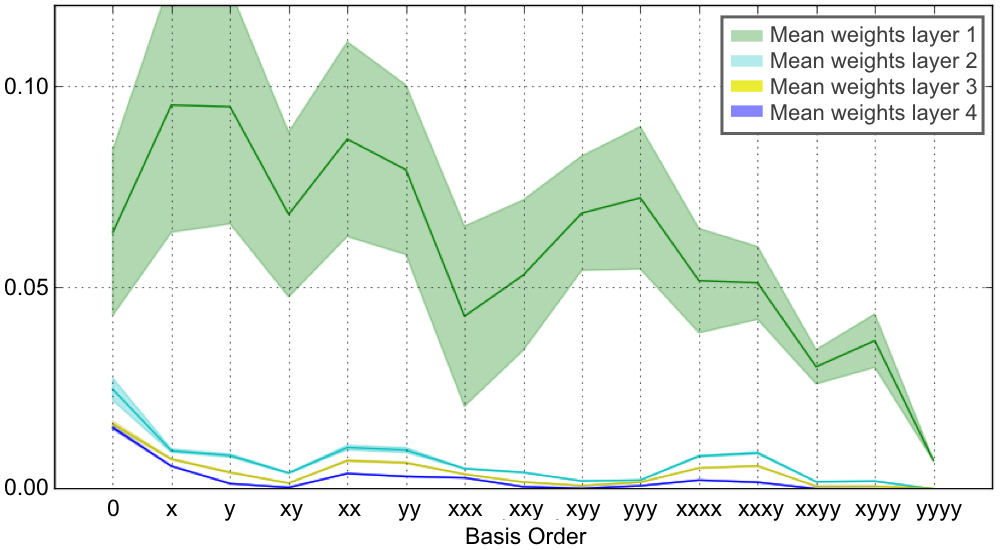}
\vspace{-5mm}
\caption{Mean of filter weights and variances per layer for 15 basis filters with no scale, as trained on ILSVRC2012-100 subset. Note that the lower order filters have the highest weights while zero-order filters are most effective in higher layers for combinations of lower responses. 
\label{fig:alphaPerLayer}}
\end{figure}

\textbf{Analysis of network layers}.
For the network RFNiN 4th-order Figure~\ref{fig:alphaPerLayer} provides an overview of the range of basis weights per effective filters in all layers, where the x-axis indexes the spatial derivative index and y-axis the mean value plus standard deviation of weights per layer over all effective filter kernels. The figure indicates that weights decrease towards higher orders as expected. Furthermore zero order filters have relatively high weights in higher layers, which hints to passing on scaled incoming features.

\textbf{Comparison to Network in Network}.
The champion RFNiN in table~\ref{tab:ILSVRC100} slightly outperforms the Network in Network with the same setting and training circumstances while only having 94 instead 520,000 spatial filters in the network in total. Note that the number of parameters is relatively similar though, as the scale component increases the number of basis functions per filter significantly. The result shows that our basis representation is sufficient for complex tasks like Imagenet. 

\textbf{Refactorize Network in Network}.
To illustrate that our proposed model is not merely a change in architecture we compare to a third architecture. We remove the Gaussian basis and we re-factorize the NiN such that it becomes identical to RFNiN. Both have almost the same number of parameters, but the NiN-factorize has a freely learnable basis. Re-factorizing only the first layer and leaving the rest of the network as in the original NiN, in table~\ref{tab:refactor} we show that a Gaussian basis is superior to a learned basis. When re-factorizing all layers, RFNiN-Scale $3^\text{rd}$-order results are superior by far to the identical NiN-factorize All Layers.

\begin{table}
\centering
\begin{tabular}{llll} \toprule 
Model&Basis&$\#$Params&Top-1\\ \midrule 
NiN-refactor Layer 1&Free&7.47M&64.10\%\\
RFNiN-refactor Layer 1&Gauss&7.47M&68.63\%\\ \midrule
NiN-refactor All Layers&Free&6.87M&38.02\%\\
RFNiN-Scale $3^\text{rd}$-order&Gauss&6.83M&69.65\%\\  \bottomrule
\end{tabular}
\caption{Classification on ILSVRC2012-100 to illustrate influence of factorization on performance. The results show that the advantage of the Gaussian basis is substantial and our results are not merely due to a change in architecture.}
\label{tab:refactor}
\end{table}

\subsection{Experiment 2: Scattering and RFNNs}

\textbf{Small simple domain}. We compare an RFNN to Scattering in classification on reduced training sizes of the MNIST dataset. This is the domain where Scattering outperforms standard CNNs~\cite{brunaTPAMI13scattering}. We reduce the number of training samples when training on MNIST as done in~\cite{brunaTPAMI13scattering}.
The network architecture and training parameters used in this section are the same as in~\cite{zeiler2013stochastic}. The RFNN contains 3 layers with a third order basis on one scale as a multiscale basis didn't provide any gain. Scale and order are determined on a validation set. Each basis layer is followed by a layer of $\alpha_N=64$ 1x1 units that linearly re-combine the basis filters outcomes. As comparison we re-implement the same model as a plain CNN. The CNN and Scattering results on the task are taken from~\cite{brunaTPAMI13scattering,ranzato2007unsupervised}.

% and plot it in the same figure, to assure that our performance is not due to the model architecture itself

\begin{figure}
\includegraphics[width=0.5\textwidth, center]{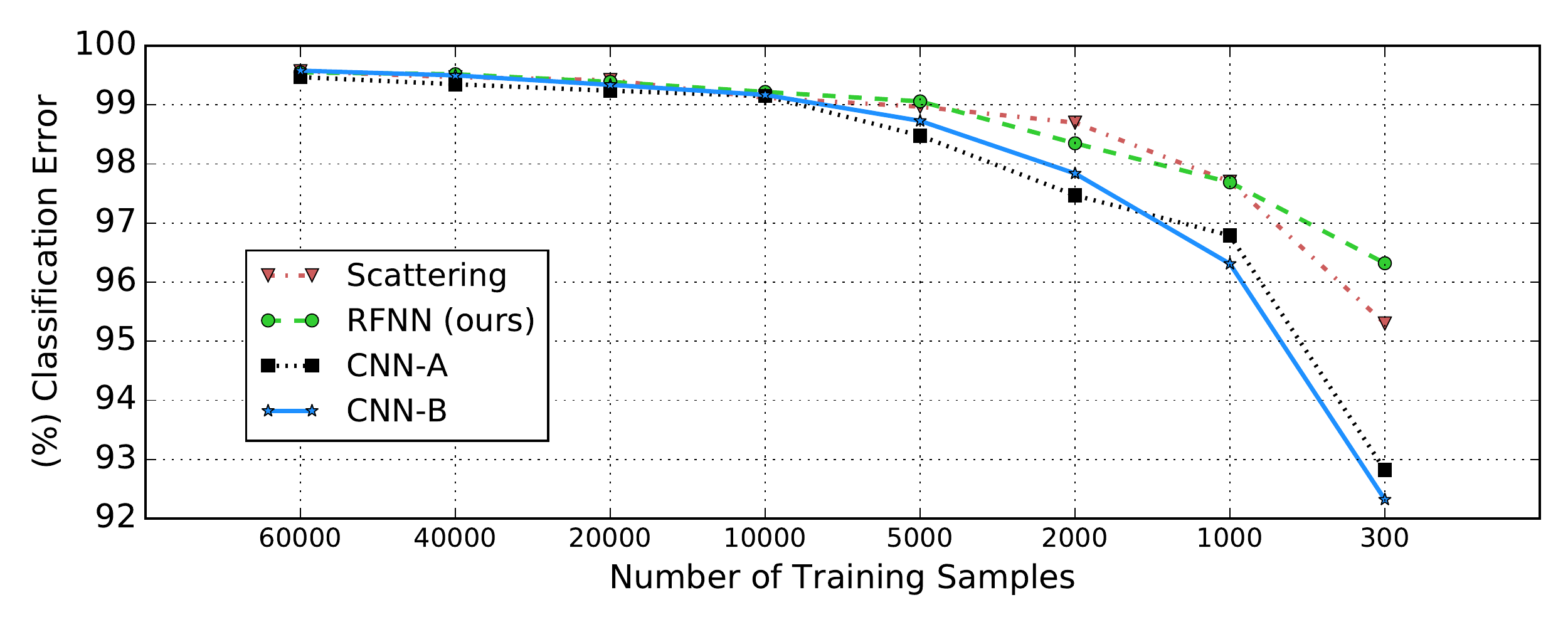}
\caption{Classification performance of the Scattering Network on various subsets of the MNIST dataset. In comparison the state of the art CNN-A from~\cite{ranzato2007unsupervised}. RFNN denotes our receptive field network, with the same architecture as CNN-B. Both are shown, to illustrate that good performance of the RFNN is not due to the CNN architecture, but due to RFNN decomposition. Our RFNN performs on par with Scattering, substantially outperforming both CNNs.}
\label{fig:MNIST}
\end{figure}

Results are shown in Figure~\ref{fig:MNIST}, each number is averaged over 3 runs. For the experiment on MNIST the gap between the CNNs and networks with pre-defined filters increases when training data is reduced, while RFNN and Scattering perform on par even at the smallest sample size.
\textbf{Large complex domain}.
We compare against Scattering on the Cifar-10 and Cifar-100 datasets, as reported by the recently introduced Deep Roto-Translation Scattering approach~\cite{oyallon2015deep}, a powerful variant of Scattering networks explicitly modeling invariance under the action of small rotations. This is a domain where CNNs excel and learning of complex image variabilities is key.

The RFNiN is again a variant of the standard NiN for Cifar-10. It is similar to the model in experiment 1, just that it has one basis layer, two 1x1 convolution layers and one pooling layer less and the units in the 1x1 convolution layers are 192 in the whole network. Furthermore, we show performance of the state-of-the-art recurrent convolutional networks~(RCNNs)~\cite{liang2015recurrent} for comparison. 

The results in Table~\ref{tab:SCAT} show a considerable improvement on Cifar-10 and Cifar-100 when comparing RFNiN to Roto-Translation Scattering \cite{oyallon2015deep}, which was designed specifically for this dataset. RCNNs performance is considerably higher as they follow a different approach to which structured receptive fields can also be applied if desired. \\

\textbf{RFNNs are robust to dataset size}.
From these experiments, we conclude that RFNNs combine the best of both worlds. We outperform CNNs and compete with Scattering when training data is limited as exemplified on subsets of MNIST. We capture complex image variabilities beyond the capabilities of Scattering representations as exemplified on the datasets Cifar-10 and Cifar-100 despite operating in a similarly smooth parameter space on a receptive field level.

\begin{table}
\centering
\begin{tabular}{lll}\toprule
Model & Cifar-10 & Cifar-100\\ \midrule
Roto-Trans Scattering & 82.30\% & 56.80\% \\
RFNiN & 86.31\% & 63.81\%\\ \midrule
%NiN (Caffe)&87.30\%&\%&-\%\\ 
%NiN~\cite{}&89.59\%&64.32\%&99.53\%\\
RCNN&91.31\%&68.25\%\\
\bottomrule
\end{tabular}
\caption{Comparison against Scattering on a large complex domain. State-of-the-art comparison is given by RCNN. RFNiN outperforms Scattering by large margins.}
\label{tab:SCAT}
\end{table}

\subsection{Experiment 3: Limiting datasize}

To demonstrate the effectiveness of the RF variant compared to the Network in Network, we reduce the number of classes in the ILSVRC2012-dataset from 1000 to 100 to 10, resulting in a reduction of the total number of images on which the network was trained from 1.2M to 130k to 13k and subsequent decrease in visual variety to learn from. To demonstrate performance is not only due to smaller number of learnable parameters, we evaluate two RFNiN versions. RFNiN-v1 is RFNiN-Scale $3^\text{rd}$-order from table\ref{tab:ILSVRC100}. RFNiN-v2 is one layer deeper and wider [128/128/384/512/1000] version of the RFNiN-v1, resulting in 3 million additional parameters, which is 2,5 million more than NiN.

The results in table~\ref{tab:param} show that compared to CNNs the RFNiN performance is better relatively speaking when the number of samples and thus the visual variety decreases. For the 13k ILSVRC2012-10 image dataset the gap between RFNiN and NiN increased to 8.0\% from 2.4\% for the 130k images in ILSVRC2012-100 while the best RFNiN is inferior to NiN by 2.98\% for the full ILSVRC2012-1000. This supports our aim that RFNiN is effectively incorporating natural image priors, yielding a better performance compared to the standard NiN when training data and variety is limited, even when having more learnable parameters. Truly large datasets seem to contain information not yet captured by our model.

\begin{table}
\begin{center}
\begin{tabular}{@{}lcccc@{}}
\toprule
Model&$\#$Params&1000-class&100-class&10-class\\\midrule 
%&&(new)&(paper)&(paper)\\ \midrule 
NiN&7.5M&56.78\%&67.30\%&76.97\%\\
RFNiN-v1&6.8M&50.08\%&69.65\%&85.00\%\\
RFNiN-v2&10M&54.04\%&70.78\%&83.36\%\\
%\midrule
%GoogLenet&xx&xx&xx&xx\\
\bottomrule
\end{tabular}
\caption{Three classification experiments on ILSVRC2012 subsets. Results show that the bigger model (RFNiN-v2) performs better than RFNiN-Scale $3^\text{rd}$-order (RFNiN-v1) on the 1000-classes while on 100-class and 10-class, v1 and v2 perform similar. The gap between RFNiN and NiN increases for fewer classes.}
\label{tab:param}
\end{center}
\end{table}

% The second part of the experiments is concerned with comparison to the scattering approach to show, that we outperform it on very small MNIST subsets as an example of very small non-natural image datasets where the scattering network has been shown to be state of the art~\cite{brunaTPAMI13scattering}, while we also perform better on the more challenging datasets Cifar-10/100, shown in [REF].

%We compare our approach to scattering networks on three datasets. We use multiple averaged runs on a tiny subset of MNIST to illustrate our ability to tackle small data non-natural image problems. On this problem scattering is de-facto state of the art. Furthermore we compare on two more challenging benchmarks, Cifar-10 and Cifar-100.

\subsection{Experiment 4: Small realistic data}

%We do two experiments on medical images, to show the value of our approach for images where pre-training doesn't work (2D Bone texture), or is simply not possible (3D Brain Scans).

We apply an RFNiN to 3D brain MRI classification for Alzheimer's disease~\cite{cuingnet2011automatic} on two popular datasets. Neuroimaging is a domain where training data is notoriously small and high dimensional and no truly large open access databases in a similar domain exist for pre-training. 

We use a 3-layer RFNiN with filters sizes [128,96,96] with a third order basis in 3 scales $\sigma \in \{1,4,16\}$. This time wider spaced, as the brains are very big objects and are centered due to normalization to MNI space with the FSL library~\cite{jenkinson2012fsl}. Each basis layer is followed by one 1x1 convolution layer. Global average pooling is applied to the final feature maps. The network is implemented in Theano~\cite{bastien2012theano} and trained with Adam~\cite{kingma2014adam}.

The results are shown in table~\ref{tab:brains}. Note that~\cite{gupta2013natural,DBLP:journals/corr/PayanM15} train on their own subset and use an order of magnitude more training data. We follow standard practice~\cite{cuingnet2011automatic} and train on a smaller subset. Nevertheless we outperform all published methods on the ADNI dataset. The same 3 layer NiN as our RFNiN model has $84.21\%$ accuracy, more than $10\%$ worse while being hard to train due to unstable convergence. On the OASIS AD-126 Alzheimer's dataset~\cite{marcus2007open}, we achieve an accuracy of $80.26\%$, compared to $74.10\%$ with a SIFT-based approach~\cite{chen2014detecting}. Thus, we show our RFNiN can effectively learn comparably deep representations even when data is scarce and exhibits stable convergence properties.

\begin{table}
\begin{center}
\begin{tabular}{@{}llll@{}}\toprule
3D MRI classification&Accuracy&TPR&SPC\\ \midrule
%3D-RFNN A&99.32\%& &\\
3D-RFNiN (ours)&\textbf{97.79\%}& \textbf{97.14\%}& \textbf{98.78\%} \\
%3D-CNN A&98.79\%& &\\
%3D-CNN B&(Chance)\%& &\\
ICA~\cite{yang2011independent}&80.70\%&81.90\%&79.50\%\\
Voxel-Direct-D-gm~\cite{cuingnet2011automatic}&-&81.00\%&95.00\%\\ \midrule
3D-CNN~\cite{DBLP:journals/corr/PayanM15}&95.70\%&-&-\\
NIB~\cite{gupta2013natural}&94.74\%&95.24\%&94.26\%\\
\bottomrule
\end{tabular}
\end{center}
\vspace{-5mm}
\caption{Alzheimer's classification with 150 train and test 3D MRI images from the widely used ADNI benchmark. RFNiN, ICA and Voxel-Direct-D-gm are trained on the subset introduced in~\cite{cuingnet2011automatic}, 3D-CNN and NIB were trained on their own subset of ADNI, using an order of magnitude more training data. RFNiN outperforms all published results. Reported is accuracy, true positive rate and specificity.}
\label{tab:brains}
\end{table}

%% file: sections/discussion.tex
\section{Discussion}
The experiments show that structuring convolutional layers with a filter basis grounded on Scale-space principles improves performance when data is limited. The filter basis provides regularization especially suited for image data by restricting the parameter space to smooth features up to fourth order. The Gaussian derivative basis opens up a new perspective for reasoning in CNNs, connecting them with a rich body of prior multiscale image analysis research that can now be readily incorporated into the models. This is especially interesting for applications where model insight and control is key. 

We  illustrated the effectiveness of RFNNs on multiple subsets of Imagenet, Cifar-10, Cifar-100 and MNIST. The choice of a third order Gaussian basis is sufficient to tackle all datasets which is in accordance with prior research~\cite{koenderinkBioCyb87localGeoInVisSys,brunaTPAMI13scattering}. While it remains an open problem to match the performance of CNNs on very large datasets like the 1000-class ILSVRC2012, our results show that the RFNN method outperforms CNNs by large margins when data are scarce. It can also outperform CNNs on challenging medium sized datasets while being superior to Scattering on large datasets despite having more parameters as the pre-defined basis restriction allows the network to devote its full capacity to a sensible feature spaces. As a small data real world example, we verify our claims with 3D MRI Alzheimer's disease classification on two datasets where we consistently achieve competitive performance including the best results on the widely used ADNI dataset.

\textbf{Acknowledgements.} We thank Rein van den Boomgaard and Silvia-Laura Pintea for insightful discussions. This work has been conducted with data from the Alzheimer's Disease Neuroimaging Initiative (ADNI) and the Open Access Series of Imaging Studies (OASIS).